\documentclass[letterpaper]{article}

\usepackage{natbib}

\usepackage{alifeconf}  

\usepackage{booktabs} 
\usepackage{amssymb}
\usepackage[ruled]{algorithm2e}
\usepackage{courier}
\usepackage[colorinlistoftodos]{todonotes}
\usepackage{subcaption}

\title{Safe Reinforcement Learning through Meta-learned Instincts}
\author{Djordje Grbic and Sebastian Risi\\
\mbox{}\\
IT University of Copenhagen, Denmark\\ \{djgr, sebr\}@itu.dk } 

\begin{document}
\maketitle

\begin{abstract}

An important goal in reinforcement learning is to create agents that can quickly adapt to new goals while avoiding situations that might cause damage to themselves or their environments. 
One way agents learn is through exploration mechanisms, which are needed to discover new policies. However, in deep reinforcement learning, exploration is normally done by injecting noise in the action space. While performing well in many domains, this setup has the inherent risk that the noisy actions performed by the agent lead to  unsafe states in the environment. Here we introduce a novel approach called \emph{Meta-Learned Instinctual Networks} (MLIN) that  allows  agents to safely learn during their lifetime  while avoiding potentially hazardous states. At the core of the approach is a plastic network trained through reinforcement learning and an evolved  ``instinctual'' network, which does not change during the agent's lifetime but can modulate the noisy output of the plastic network. We test our idea on a simple 2D navigation task with no-go zones, in which the agent has to learn to approach new targets during deployment. MLIN outperforms standard meta-trained networks and allows agents to learn to navigate to new targets without colliding with any of the no-go zones. These results suggest that meta-learning augmented with an instinctual network is a  promising new  approach for RL in safety-critical domains.
\end{abstract}

\section{Introduction}

\begin{figure}[htb]
\centering
\includegraphics[width=0.3\textwidth]{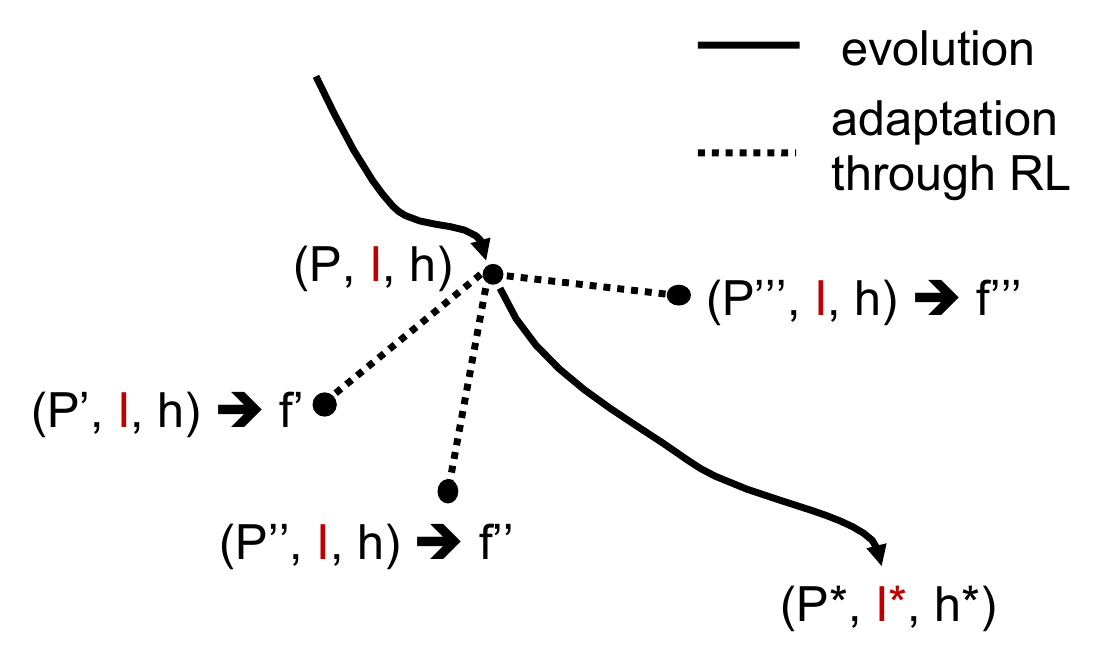}
\caption{\textbf{Meta-Learned Instinctual Networks (MLIN).} \normalfont Evolution determines the initial parameters $P$ of a policy network together with hyperparameters $h$. During an evaluation that includes adaptation to different tasks (three in this example), policy parameters $P$ are updated through RL, and performance is evaluated with the  updated policy network parameters, returning a fitness for each task. Modified policy parameters are not inherited by the next generation. The novel addition in this paper is the parameters $I$ of an instinctual network, which can suppress the activation of the policy network in hazardous situations, which is also evolved. Importantly, the parameters $I$ stay constant during the RL adaptation process (i.e.\ during the agent's lifetime). To generate the next generation, the initial parameters of $P$, hyperparameters $h$, and weights of the instinctual network $I$ are mutated to produce (P*, I*, h*).} 
\label{fig:overview}
\end{figure}

While especially deep reinforcement learning (RL) approaches have shown impressive results across a large variety of different domains  \citep{justesen2019deep, li2017deep}, creating  RL approaches that respect safety concern has been recognized as a major challenge  \citep{raybenchmarking,wainwright2019safelife,ortega2018building}. Reinforcement learning, in particular, is based on the idea of learning through exploration, in other words: trial and error. However, trying out different options in an environment without any restrictions can be inherently risky. The agent might try behaviors that lead to catastrophic outcomes from which recovery or further learning is impossible. While this is not necessarily a problem in simulated environments, it becomes a more challenging issue if we would like these systems to someday work well in the real world. For example, a factory robot can not just randomly try out actions but has to make sure that the options tried do not pose any danger to humans working alongside such systems. 

In contrast to current RL approaches \citep{kenton2019generalizing}, animals in nature developed efficient strategies that often prevent them from trying out actions that are potentially dangerous to their lives. In particular, animals and humans possess many different \emph{instincts}, which are innate behaviors provided by evolution that are not modified through lifetime learning. For example, six-month-old infants have a congenital fear of spiders and snakes \citep{hoehl2017itsy}, likely because this evolved instinctual fear improved our chances of survival. Rats instinctively and without any learning avoid a specific compound found in the urine of carnivores  \citep{ferrero2011detection}, which triggers an avoidance behavioral response. 

The main idea in the approach introduced in this paper is to allow agents to evolve similar innate capabilities that help them to avoid potentially dangerous situations. The novel approach, called \emph{Meta-Learned Instinctual Networks} (MLIN), builds on ideas from training agents for fast adaptation through meta-learning  
\citep{finn2017model,fernando2018meta,grbicRisi2019}. A novel insight in this paper is that \emph{meta-learning can be an effective approach for AI safety} by jointly evolving the initial parameters of a policy network that can adapt quickly during deployment through RL \citep{fernando2018meta}, with the weights of an instinctual network that only changes during evolution and can modulate the noisy actions of the policy network to prevent the agent from encountering potentially dangerous situations (Fig.~\ref{fig:overview}). Importantly, once the evolutionary training is done, safe and fast adaptation to new goals is still possible through RL.

The results in a simple 2D navigation domain demonstrate that an instinctual network is critical to allowing an agent to learn to navigate to different target areas \emph{during its lifetime} while avoiding hazardous areas in the environment. In the future, the idea of combining meta-learning with an instinctual network could now enable safer forms of AI across a range of different tasks.

\section{Background}
This section reviews policy gradient methods, which allow the agents to adapt during their lifetime, and related work on meta-learning.

\subsection{Policy Gradient Methods}
In reinforcement learning (RL) an agent is tasked with maximizing some reward by interacting with its environment. The agent follows a policy, which takes an observed environment state and returns an action to the environment. The environment is often modeled by an initial state distribution, a state transition distribution, and a reward function. From an initial to a termination state, an agent goes through a sequence of actions called episode or trajectory. The agent tries to optimize its performance with respect to the cumulative reward collected through an episode. Typically, an agent has to sample trajectories exploring the environment that will help it optimize the performance. 

Policy gradient methods \citep{Williams1992}, which we employ in our experiments, are a set of methods that optimize parameterized policies with respect to expected episode returns (sum of discounted episode rewards) by a gradient-based optimization algorithm. We denote a parameterized policy with $f_\theta(action|state)$, where $\theta$ are the parameters of the policy. The parameterized policies define a distribution of actions contingent on the current state and policy parameters. The methods compute an estimator of the policy gradient and return it to a gradient-based optimization algorithm. 
The general equation for the policy gradient calculation is: 
\begin{equation}
    \nabla_\theta J(\theta) = \mathbb{E}_\theta [\sum_{t=1}^T\nabla_\theta \log f_\theta(a_t|s_t)R_c(s_t)], 
    \label{eq:expPolicyGrad}
\end{equation}
where $T$ is the total number of steps over all trajectories, and $R_c(s_t)$ is the estimated return of the state $s_t$. In the formula, the expectation $\mathbb{E}_\theta$ is approximated with a finite batch of sampled trajectories. 

Since vanilla policy gradient methods \citep{Williams1992} are prone to catastrophically large policy updates, PPO \citep{schulman2017proximal} became a popular upgrade on the base version of the method. PPO is a simplified version of the TRPO algorithm \citep{schulman2015trust} which limits the policy update to a "trust region" to prevent the learning instabilities and catastrophic fall in the performance.

\subsection{Meta-learning}
Creating agents that can adapt quickly is one of the long-term goals in AI research. While current deep learning systems are good at learning a particular task, they still struggle to learn new tasks quickly; meta-learning tries to address this challenge. The idea of meta-learning or \emph{learning to learn} (i.e.\ learning the learning algorithms themselves) has been around since the late 1980s and early 1990s \citep{schmidhuber1992learning,schmidhuber1993neural} and is now a very active area of research.

A recent trend in meta-learning, which we follow in this paper, is to find good initial weights in an outer loop from which adaptation to different tasks can be performed in a few iterations in an inner optimization loop. This approach was first introduced by \citet{finn2017model} and is called Model-Agnostic Meta-Learning (MAML). In the approach presented in this paper, we use an evolutionary meta-learning variant, in which evolution is trying to find good initial neural network parameters that allow an inner RL loop to adapt quickly \citep{fernando2018meta, grbicRisi2019}.

A less explored meta-learning area  is the evolution of plastic networks that change at various timescales through local learning rules, such as Hebbian learning. These evolving plastic ANNs (EPANNs) are motivated by the promise of discovering principles of neural adaptation, learning, and memory \citep{soltoggio2017born}. While the paper presented here does not deal with neural networks that can learn through local learning rules, adding such learning to our system is an interesting future extension. 

While the above mentioned meta-learning approaches allow agents to adapt faster, they do not take into account any safety concerns while learning. We will review existing approaches for safer AI in the next section.

\subsection{AI Safety}
In this paper, we focus on safety in the context of deep reinforcement learning approaches. For a broader overview of work in AI safety, we refer the interested reader to the reviews by \citet{pecka2014safe} and \citet{garcia2015comprehensive}. Most work in this area focuses on \emph{constrained RL} 
\citep{altman1999constrained,wen2018constrained}. In constrained RL, safety requirements are formulated as constraints, which are states and behaviors the system should avoid. These constraints are often incorporated into RL algorithms through reward functions. However, it is not always clear what the optimal weighting between the actual task reward and the penalty for violating the constraints should be. For example, if the penalty is chosen too small, the agent will learn unsafe actions, while it might not learn anything at all if the penalty is too high \citep{raybenchmarking,pham2018optlayer,achiam2017constrained,dalal2018safe}.

Another approach to safer RL was introduced by \citet{alshiekh2018safe}, in which a reactive system, called "shield", monitors the agent's actions and corrects the actions if they would violate the pre-specified safety constraints. However, this approach relies on temporal logic specifications of the safety constraints.  Other approaches to safe deep RL include estimating the safety of trajectories through Gaussian process estimation \citep{fan2019safety}, or  reducing catastrophic events through ensembles of neural models that capture uncertainty and classifiers  trained to recognize dangerous actions \citep{kenton2019generalizing}. 

A related approach to the one introduced in this paper is called \emph{intrinsic fear} \citep{lipton2016combating}. This approach involves a second module that is trained in a supervised way to predict the probability of imminent catastrophic events, which is then integrated into a Q-learning objective. The approach presented in our paper is different in that it formulates safe learning in the context of meta-learning. During meta-training, safety violates are slowly reduced, allowing safe task adaptation after meta-training. 

\section{Approach: Meta-Learned Instinctual Networks}
The goal of the approach presented in this paper is to allow agents to learn to adapt to a variety of different tasks during their lifetime while avoiding hazardous and unsafe states in the environment. Here, 
we assume that the set of hazardous states $S_h \subset S$, where $S$ is the set of all states, is the same for all tasks the agent needs to adapt to during its lifetime.
We also assume that the undesirability of a hazardous state is communicated by sending a negative reward to the agent once reaching such a state. For example, imagine a maze with crevasses that can damage a robot, in which the robot needs to locate a goal; ideally, the robot would be equipped with a mechanism to suppresses noisy exploratory actions near crevasses.

More formally, a particular task $\mathcal{T}_i$ the robot should adapt to is sampled from a task distribution $p(\mathcal{T})$, which contains the state transition distribution $q_i(s_{t+1}| s_t, a_t)$, initial state distribution $q_i$, and the reward function $R_i(s)$. The associated functions make the task a Markov decision process with horizon $H$. Here, we define the hazard neighborhood as a set of states from which there is a non-zero probability that some available sequence of actions would lead to a hazardous state; $\{s_n \in \{S\} | \exists a_n \rightarrow q(s_{n+1}|s_n, a_n)q(s_{n+2}|s_{n+1}, a_{n+1}) ... q(s_h|s_{n+m}, a_{n+i})>0\}$, where $s_h \in S_h$.

The agent should be able to maximize the cumulative episode reward $R_c = \sum_{j=1}^H R(s_j)$, by sampling several trajectories, while minimizing the punishment for visiting hazardous zones during the trajectory sampling. To achieve this, the agent needs to know when it finds itself in a hazard's neighborhood and can thus learn to suppress unsafe exploratory actions.

\subsection{Model architecture}
\begin{figure}
\includegraphics[width=0.4\textwidth]{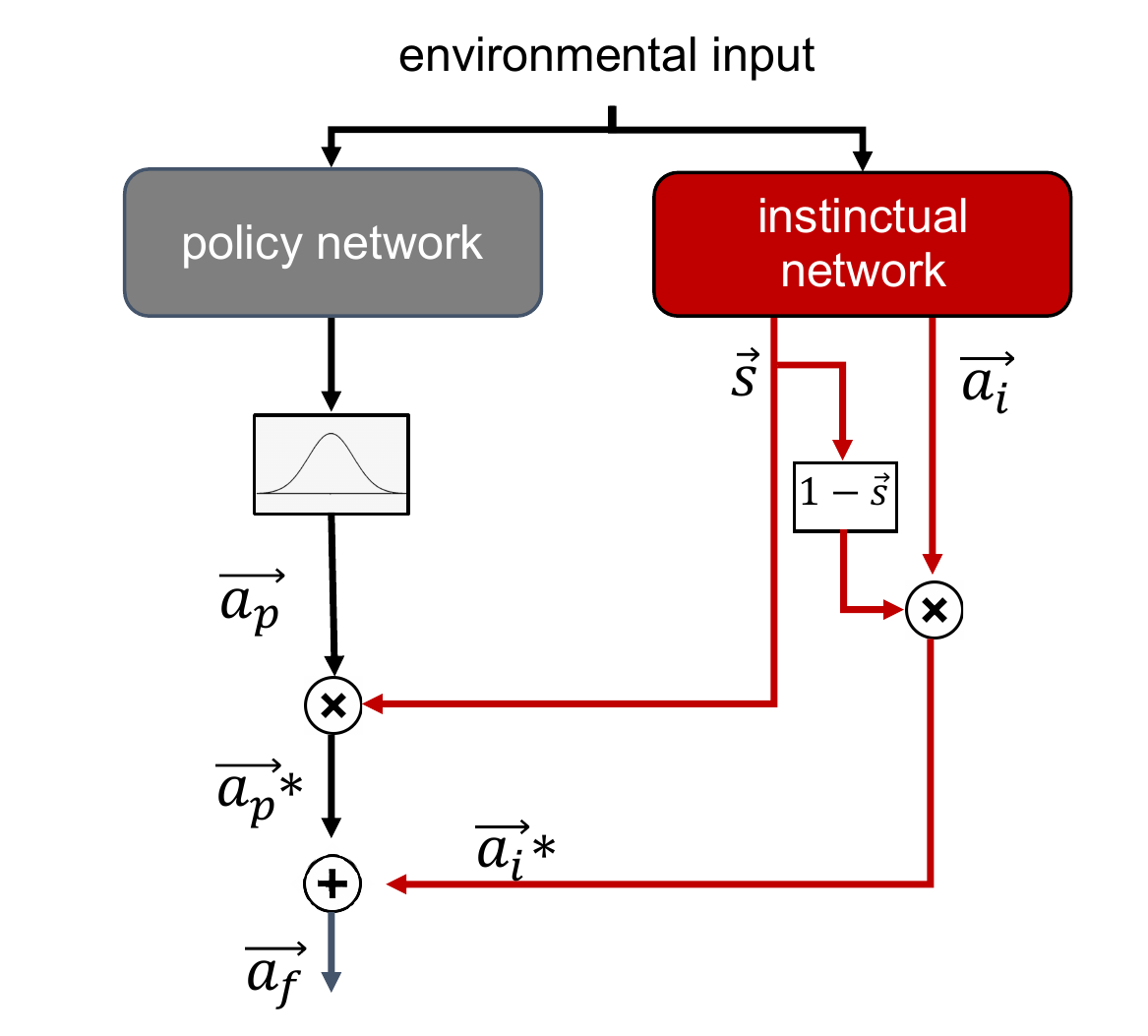}
\caption{The topology of the policy network with instinct module. \normalfont Both networks receive the same input from the environment. The instinctual network outputs an instinctual action $\vec{a_i}$ and a suppression signal $\vec{s}$. The suppression signal is a vector of values between 0 and 1 that determines the magnitude of instinctual action that will be mixed in the policy action. The suppression signal $\vec{s}$ is multiplied with policy action $\vec{a_p}$ and the opposite suppression signal $\vec{1}-\vec{s}$ is multiplied with the instinctual action $\vec{a_i}$. Two action values are finally added resulting in the final action $\vec{a_{f}}$.}
\label{fig:topology}
\end{figure}

The model architecture introduced in this paper consists of two neural network modules: a policy network and an instinctual network (Fig.~\ref{fig:topology}). The policy network is a neural network module that is trained to solve a specific task through reinforcement learning, while the instinctual network is kept fixed during task adaptation. 
The goal of the instinctual network is to safeguard the policy network from potentially dangerous actions during exploration, by modulating its outputs. The specific architecture described here is suitable for reinforcement learning problems with continuous action spaces. 

The policy network has an output for all the actions the agents can perform in the environment (e.g.\ two actions for moving in two dimensions). The instinct network has two different types of  outputs: The first output is a suppression signal and the second output is an instinctual action. Following the standard way of exploration in RL \citep{Williams1992}, the actions of the policy network $p$ are noisy; the policy network outputs a mean action $a_\mu$ that is given to a distribution (usually the normal distribution) from which the output action is sampled: $a_p^n \sim \mathcal{N}(a_\mu^n, \sigma^n)$, where $\sigma$ is part of the policy parameters $\theta^p$, and $n$ denotes $n^{th}$ action dimension. The suppression signal $\vec{s}$ from the instinct module is multiplied with the action vector generated by the policy network. The suppression signal has the same number of dimensions as the action vector, such that each dimension can be suppressed separately. Another vector is created by subtracting the suppression signal values from one to create the suppression signal that will modulate the instinctual action $\vec{a_i}$, where $i$ denoted the instinctual network. More precisely, the activation of the network follows the steps below:
\begin{enumerate}
    \item instinct network outputs two vectors, $\vec{s}$ and $\vec{a_i}$, where $\vec{s}$ elements are in the interval $[0, 1]$; 
    \item policy networks outputs $\vec{a_p}$;
    \item $\vec{a_p}$ gets modulated with the suppression vector, $\vec{a_p^*} = \vec{s} \odot \vec{a_p}$, where $\odot$ is the element-wise multiplication of vectors;
    \item $\vec{a_i^*} = \vec{a_i} \odot (\vec{1} - \vec{s})$;
    \item final action vector $\vec{a_{f}}$ is the sum of two modulated action vectors, $\vec{a_{f}} = \vec{a_p^*} + \vec{a_i^*}$.
\end{enumerate}

\subsection{Meta-training}
The question here is how to train an instinctual network that keeps the agent out of harm's way together with a policy network that should be able to adapt quickly to new goals. One of the main insights in the work presented here is that we can use an evolutionary meta-learning approach \citep{fernando2018meta,grbicRisi2019} to train a policy that can adapt quickly and \emph{safely} to different tasks. The whole training procedure runs two training loops: an evolutionary outer loop, and a task-adaptation inner loop (Alg.\ref{alg:meta} and Fig.\ref{fig:overview}). 

In the outer evolutionary loop, a simple genetic algorithm (GA)  is optimizing the initial parameters of the policy network (weights and  Gaussian action noise $\sigma$), the weights of the instinctual network, and a learning rate used by the RL algorithm in the inner loop. The innovation in this paper is the instinctual neural network, whose weights are only updated through mutations during the outer loop and are not modified in the inner loop. In other words, instincts are not modified during an agent's lifetime.

The evolved parameters of a particular genome $g$ passed to the inner loop are the parameters of the policy network $\theta_g^p$, the parameters of the instinctual network $\theta_g^i$, and a learning rate $\alpha_g$. The inner loop takes the evolved parameters and evaluates their performance by cycling through tasks $\mathcal{T}_i \in \mathcal{P}_\mathcal{T}$. For each task $\mathcal{T}_i$, the inner loop collects trajectories $(s_0, a_0, ...,s_H)$ by sampling noisy action, produced by the policy network, modulated with the instinctual network. When the agent reaches a goal or collects $H$ state-action pairs (set to 20 in this paper), it is repositioned back to the center and the new trajectory begins.  This process is repeated until the agent collects  2,000 state-action pairs $(s, a)$ for each task $\mathcal{T}_i$.
The algorithm collects the sum of safety violation punishments $V$ over the sampled trajectories. Using the collected data from the trajectories, a gradient-based optimization algorithm modifies the weight values of the policy network: $\theta^p = \theta^p_g - \alpha \nabla \mathcal{L}(f_{\theta^p})$. 

Our specific implementation uses PPO \citep{schulman2017proximal} for the policy gradient calculation $\nabla \mathcal{L}(f_{\theta^p})$, and the Adam optimizer \citep{kingma2014adam} for the gradient update of the policy network. The PPO algorithm takes the action log-probabilities ($\log f_\theta(s, \vec{a_p})$) sampled from a distribution defined by the policy network (Eq.~\ref{eq:policyGrad}), not the instinct-modulated actions that are given to the environment.

After the update, the algorithm samples the final trajectory where the policy network generates actions by taking the mean $a_\mu$ action of the $f_\theta(\cdot)$ distribution. The cumulative episode reward is added to the hazard violation punishments ($episode\_reward(\theta^p,\textcolor{red}{\theta^i}) + \textcolor{red}{V}$).   Hazard violation punishment is based on how often the agent enters one of the undesired states. Note that the weight values of the policy network after the gradient-based update are discarded after each task ($\theta^p$ $\leftarrow$ $\theta_g^p$). In other words, parameter adaptations to a specific task are not inherited (i.e.\ they are non-Lamarckian). The final evaluation of the evolved parameters is the sum of task evaluations $F_g$ for each task visited in the inner loop. The parameters ($\theta_g^p$, $\theta_g^i$, and $\alpha_g$) are optimized in the outer loop based on the evaluation values $F_g$.

\begin{algorithm}           
  \small
    \SetCustomAlgoRuledWidth{0.45\textwidth}  
    \caption{Meta-Learned Instinctual Networks (MLIN)}
    \ForEach{genome $g$ in population}{
        \emph{use evolved policy net parameters $\theta_g^p$, instinct net parameters $\theta_g^i$, and learning rates $\alpha$}\;
        $\theta^p$ $\leftarrow$ $\theta_g^p$,
        \textcolor{red}{$\theta^i$ $\leftarrow$ $\theta_g^i$,},
        $\alpha$ $\leftarrow$ $\alpha_g$
        $F_g \leftarrow 0$,
        \textcolor{red}{$V \leftarrow 0$\;}
        \emph{take tasks from task collection}\;
        \For{$\mathcal{T}_i$ in $\mathcal{P}_T$ }{
           \emph{run RL for n steps}\;
           \For{$t\leftarrow 0$ \KwTo $n$}{
                \emph{run trajectories, collect rewards and \textcolor{red}{\#violations}}\;
                $a_f \leftarrow f_{\theta^p, \textcolor{red}{\theta^i}}(s_t)$\;
                $s_{t+1} \leftarrow episode\_step(a_f)$\;
                \textcolor{red}{$V$ $\leftarrow$ $V + \#violations$\;}
            }
            \emph{Run gradient-based update on the policy parameters $\theta^p$}\;
            $\theta^p = \theta^p - \alpha \nabla \mathcal{L}(f_{\theta^p})$\;
            \emph{Add fitness to overall fitness}\;
            $F_g \leftarrow F_g + [episode\_reward(\theta^p,\textcolor{red}{\theta^i}) + \textcolor{red}{V}]$\;
            \emph{Reset the values of the policy network.}\;
            $\theta^p$ $\leftarrow$ $\theta_g^p$\;
        }
    }
\label{alg:meta}
\end{algorithm}

\section{Task environment}
\begin{figure}
\centering
\includegraphics[width=0.35\textwidth]{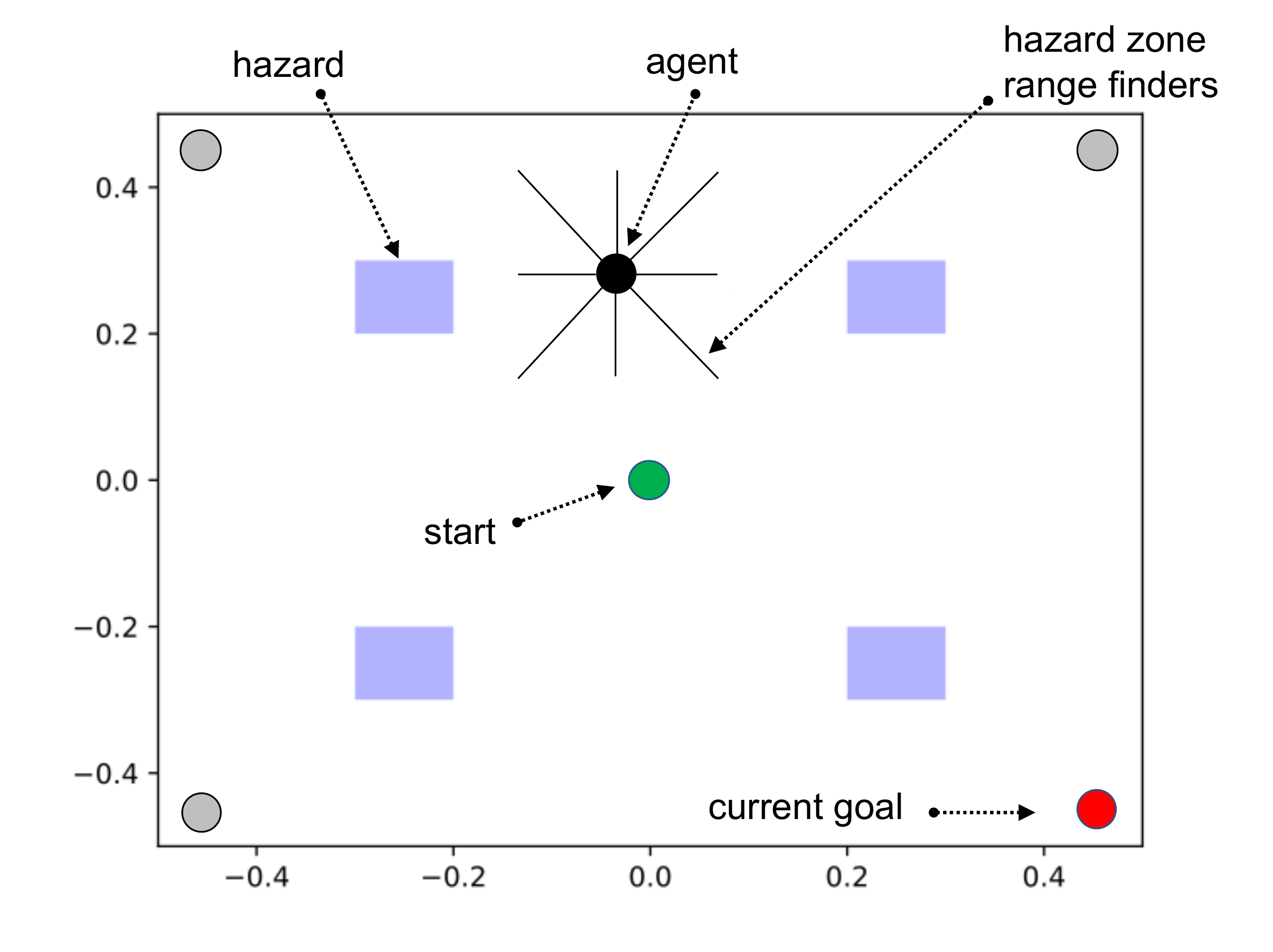}
\caption{Navigation task with hazardous areas. \normalfont The agent is spawned in the center and has to learn -- during its lifetime -- to reach a particular goal position (which it does not see). 
}
\label{fig:2D_nav}
\end{figure}

The test domain in this paper is a 2D navigation task with four hazardous areas (Fig.~\ref{fig:2D_nav}), which is inspired by the simpler 2D navigation (which does not include hazardous areas) used to evaluate the original MAML approach \cite{finn2017model}. The environment consists of an agent starting at the coordinate (0,0). The goal of the agent is to learn how to reach one of four goals $\mathcal{T}_i \in [(0.45, 0.45),$ $(0.45, -0.45),$ $(-0.45, 0.45),$ $(-0.45, -0.45)]$ only through the reward it receives at each time step. The inner loop cycles through all four goals and rewards the agent for how close it can approach them (Alg.\ref{alg:meta}). It is important to emphasize that similar to the setup of \citet{finn2017model}, the agent's neural network has no access to the location of the current goal, and the agent has to reach it only by adapting its policy through rewards. This ensures that the task indeed requires adaptation during the lifetime of the agent; if the agent could see the goal through sensors, a static policy would be able to reach each goal without having to re-adapt.

One component to reward is the negative distance of the current position to the goal state; $r_{d,t} = -\sqrt{\sum_j(T_{i,j} - s_{t,j})^2}$, where $s_t$ is the agent's current coordinate. The second component is a penalty of $r_{h,t} = -10$, which the agent receives for each step it makes in a hazard zone. The total state reward is thus calculated as: $R(s_t) = r_{d,t} + r_{h,t}$. An episode terminates if the agent gets within 0.01 units to the goal state or the episode exceeds the maximum horizon of 20 timesteps. 

The hazardous areas in the environment test the agent's ability to adapt to new goal positions in a safe way. The agent has to learn by trial to reach the goal position while avoiding the hazardous areas. The policy network and the instinctual network get the position the agent currently occupies $(x,y)$ and the eight range-finders, which detect the proximity of the hazardous areas, as input. One range-finder returns the fraction of the distance that an edge of a hazardous zone occupies. The agent outputs a movement vector $(\Delta x, \Delta y)$,  where $\Delta x$ and $\Delta y$ are in the range from -0.1 to 0.1 (Fig.~\ref{fig:2D_nav}).

\subsection{Network implementation details}
For the \emph{policy network}, we use the same architecture as in the 2D navigation task from original MAML paper \citep{finn2017model}, an actor-critic system, where actor and critic are two separate, fully connected neural networks with two hidden layers of 100 neurons each and Tanh activation function. While the task could likely be solved by a smaller network, to more easily analyze the effects of adding an instinctual network, we kept the setup as close as possible to the original MAML experiments. The policy gradient can be described as:  
\begin{equation}
    \nabla_\theta J(\theta) = E_\theta [\nabla_\theta \log f_\theta(s,a)A_\theta(s)], 
    \label{eq:policyGrad}
\end{equation}
where $A_\theta(\cdot, \cdot)$ is the advantage calculated from the critic and $f_\theta(\cdot, \cdot)$ is the output of the actor-network.
The critic-network is updated to minimize the temporal difference between predicted expected return $A_\theta(s_t)$ at state $s_t$, and the reward $R(s_t)$ updated return estimate: $A_\theta(s_t) - (R(s_t) + \gamma A_\theta(s_{t+1}))$, where $\gamma$ is the reward discount hyperparameter \citep{peters2005natural, wu2017scalable}.

 The actor outputs a mean action  for a Gaussian distribution $\mathcal{N}(\vec{a}_\mu, \vec{\sigma})$,  
 from which an action is sampled \citep{Williams1992}. During the final deterministic evaluation of the policy ($episode\_reward$ in Alg.~\ref{alg:meta}), no Gaussian noise is added. The critic outputs the predicted value (predicted future cumulative reward). The final layer of the actor-network has two outputs (Tanh) scaled to $[-0.1, 0.1]$, reflecting the 2D-navigation task action space.

The \emph{instinct module} has two hidden layers of 100 neurons, with the ReLU activation functions, and two parallel output layers (instinct action and suppression signal). Each output layer has two output neurons (2D-navigation task action space dimensions), where the suppression signal output function is the Sigmoid function, and the instinct action output function is a Tanh function with codomain scaled to $[-0.1, 0.1]$ interval.

\subsection{Optimization Details}
\label{subsection:genAlg}

The weights of both policy and instinctual networks are initialized by Kaiming uniform initialization introduced in \cite{He_2015}. Gaussian action noise parameter $\sigma$ is initialized to 0.05. A single population has 480 individuals (60 CPUs  $\times$ 8). 
Following recent trends in deep neuroevolution, we employ a simple mutation-only genetic algorithm, which has shown to rival RL methods in different domains \citep{such2017deep, risi2019deep}. In the selection step, the top 10\% best performing individuals are chosen as the parents for the following generation. Each parent makes clones, which mutate, and are placed in the next generation until the population is full. One child in the next generation is the unchanged best-performing individual from the previous generation (the elite). Similarly to the mutation operator in previous work that optimizes deep neural network \citep{risi2019deep,such2017deep}, Gaussian noise centered around the current parameter value (weight or learning rate) with an initial sd of 0.01 is added to the network's parameters. Mutation decays with a rate of 0.999, with a minimum sd of 0.001.
Each individual is evaluated on four different goals. The inner loop evaluation $F_g$ (from the previous subsection) is the genotype fitness.

We used an existing PPO implementation from \cite{pytorchrl}. The algorithm requires a set of hyperparameters that stayed constant throughout the experiments. The hyperparameters are: $\gamma$ discount factor (0.99), PPO clip parameter (0.2), PPO epoch number (4), value loss coefficient (0.5), and entropy term coefficient (0.01).

\begin{figure}
\centering
\includegraphics[width=0.4\textwidth]{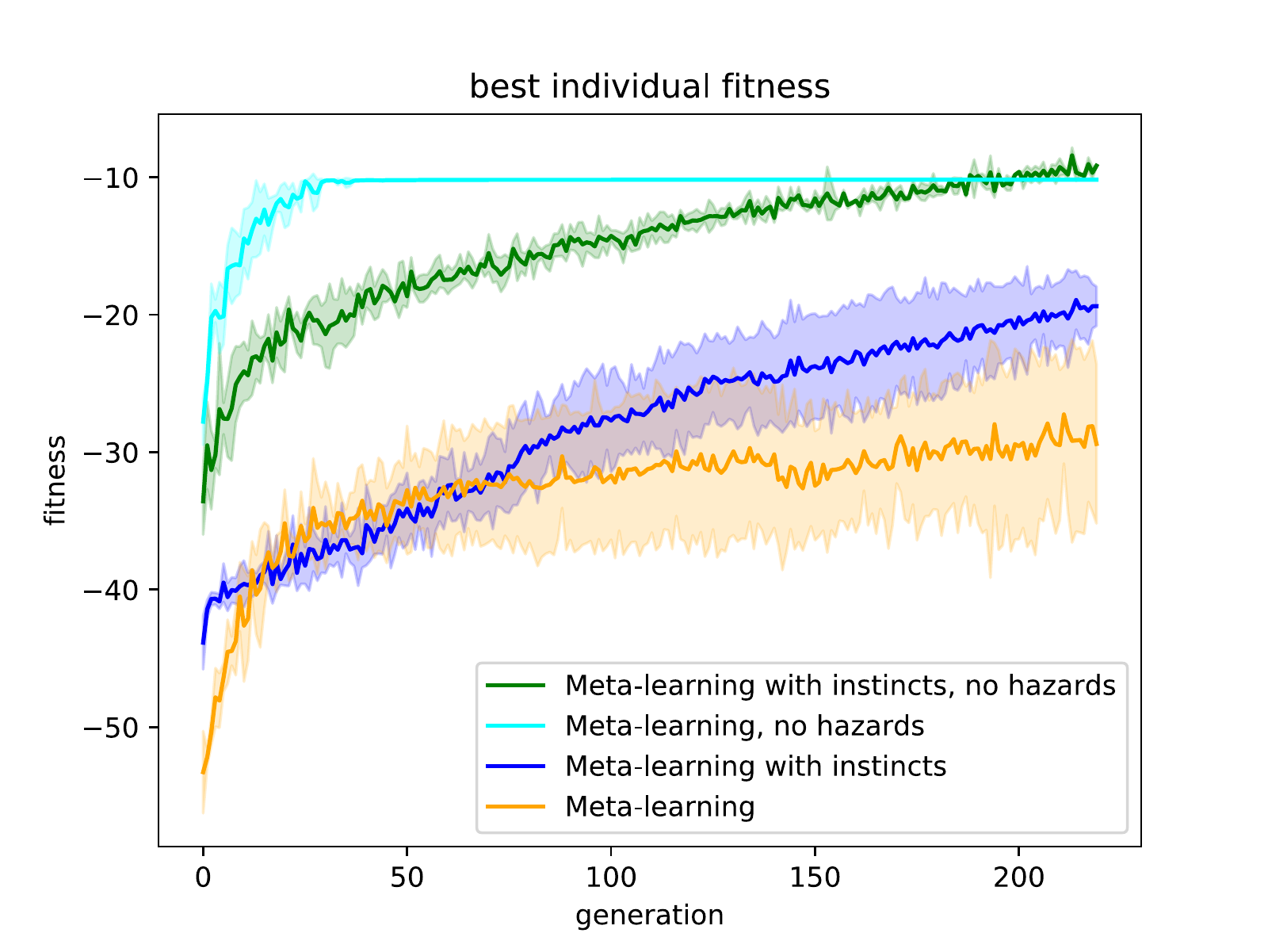}
\caption{Average fitness progress over generations. \normalfont Curves show the average fitness among five runs for the hazard and no-hazard 2D navigation task. Shaded areas show one standard deviation.}
\label{fig:fitness}
\end{figure}

\section{Results}
We compare MLIN against a meta-learning version without an instinct network in a 2D navigation task with and without hazards. For the non-instinct version, we found that scaling the output of the policy network by a factor of 0.5 gives significantly better performance and  mimics the magnitude of the average initial suppression signal in the MLIN setup.  

\begin{figure}
\includegraphics[width=0.51\textwidth]{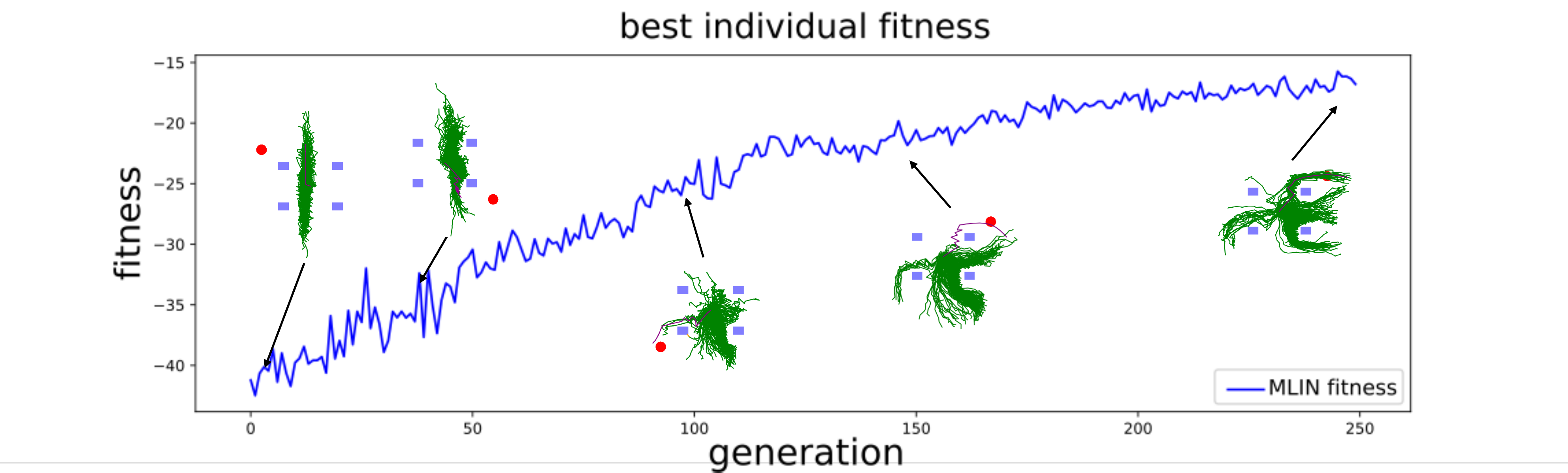}
\caption{Example evolutionary run for MLIN. 
\normalfont Shown is the fitness calculated based on the agent's distance to the goal. Early on in evolution, the agents learn to avoid the four hazardous areas but they are not able to reach any goals. The agents improve on that ability over multiple generations, after which they can safely approach the four targets.} 
\label{fig:decoupledFit}
\end{figure}

In the no-hazard environment, a meta-learning approach without instincts can quickly learn the task (Fig.~\ref{fig:fitness}), while MLIN takes longer to be optimized, likely because optimizing both an instinct and  policy network at the same time is more complicated. Final cumulative fitness is, in general, lower for the environment with hazards because the path to the goal is longer. 

However, once hazard zones are introduced, MLIN outperforms the non-instinctual version, indicating that instincts become more crucial in task that require safe learning (Fig.~\ref{fig:fitness}). A more detailed view of a particular evolutionary run is shown in Fig.~\ref{fig:decoupledFit}, which shows the progression in fitness over generations together with the exploratory behaviors performed by the best agent found so far. Since the negative reward for violating the hazard zones is an order of magnitude larger than the distance fitness (-10 for each step violating hazard zones), evolution favors models that prioritize avoiding hazards early on. 

The differences in performance of the two methods (Fig.~\ref{fig:fitness}) and their behaviors suggests that evolving only the initial parameters of a network results in a policy that learns to avoid the hazardous areas (Table~\ref{table:violations},  MLIN vs. ML \emph{without} instincts) but cannot consistently navigate to the sampled goals without colliding with the hazard zones  (Table~\ref{table:fitness}). 

\begin{figure}
\centering
\includegraphics[width=0.45\textwidth]{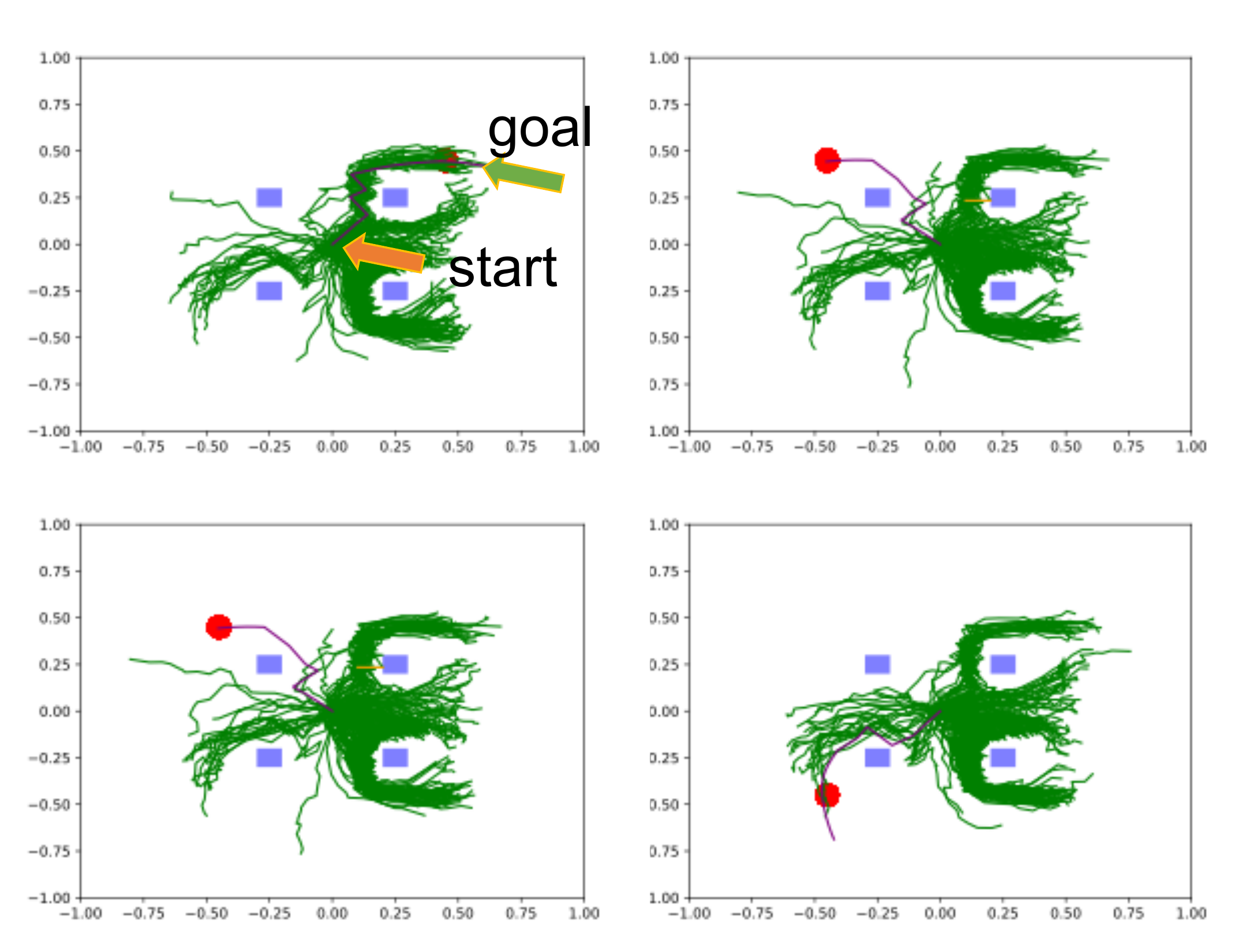}
\caption{\textbf{MLIN Goal Adaptation.} \normalfont The green lines show the exploration trajectories while the purple line is the deterministic trajectory of the model after the first gradient update. The agent is able to learn to navigate to the four target goals during its lifetime while avoiding hazard zones.}
\label{fig:trajectories}
\end{figure}

\subsection{Testing fast and safe lifetime adaptation} 

We compare how fast and safe RL can adapt an MLIN optimized network  to a new goal during its lifetime, compared to RL adapting a network with randomly initialized parameters  (pure RL without meta-training). The pure RL approach uses Kaiming weight initialization, learning rate of 0.01, and action noise $\sigma = 0.05$.

Following \citet{finn2017model}, each agent  performs $\approx$40 trajectories (4000 steps state-action samples), which were used to perform one gradient update with PPO. 
The pure RL setup reaches an average goal distance of -13.3$\pm 2.4$, while MLIN is able to adapt faster and therefore gets much closer to the four goals (-3.9$\pm$1.5) (Table~\ref{table:fitness}). Additionally, the pure RL version has an average of 75.9$\pm$48.3 safety violations while MLIN has only  0.05$\pm$0.2 (Table~\ref{table:violations}).

Fig.~\ref{fig:trajectories} shows the exploration trajectories for the best meta-trained model with  instinctual network. MLIN trained networks can consistently learn to approach the four targets while avoiding any of the hazardous zones.

\begin{figure}[ht]
        \centering
        \centering
    \begin{subfigure}{0.15\textwidth}
        \includegraphics[width=1.0\textwidth]{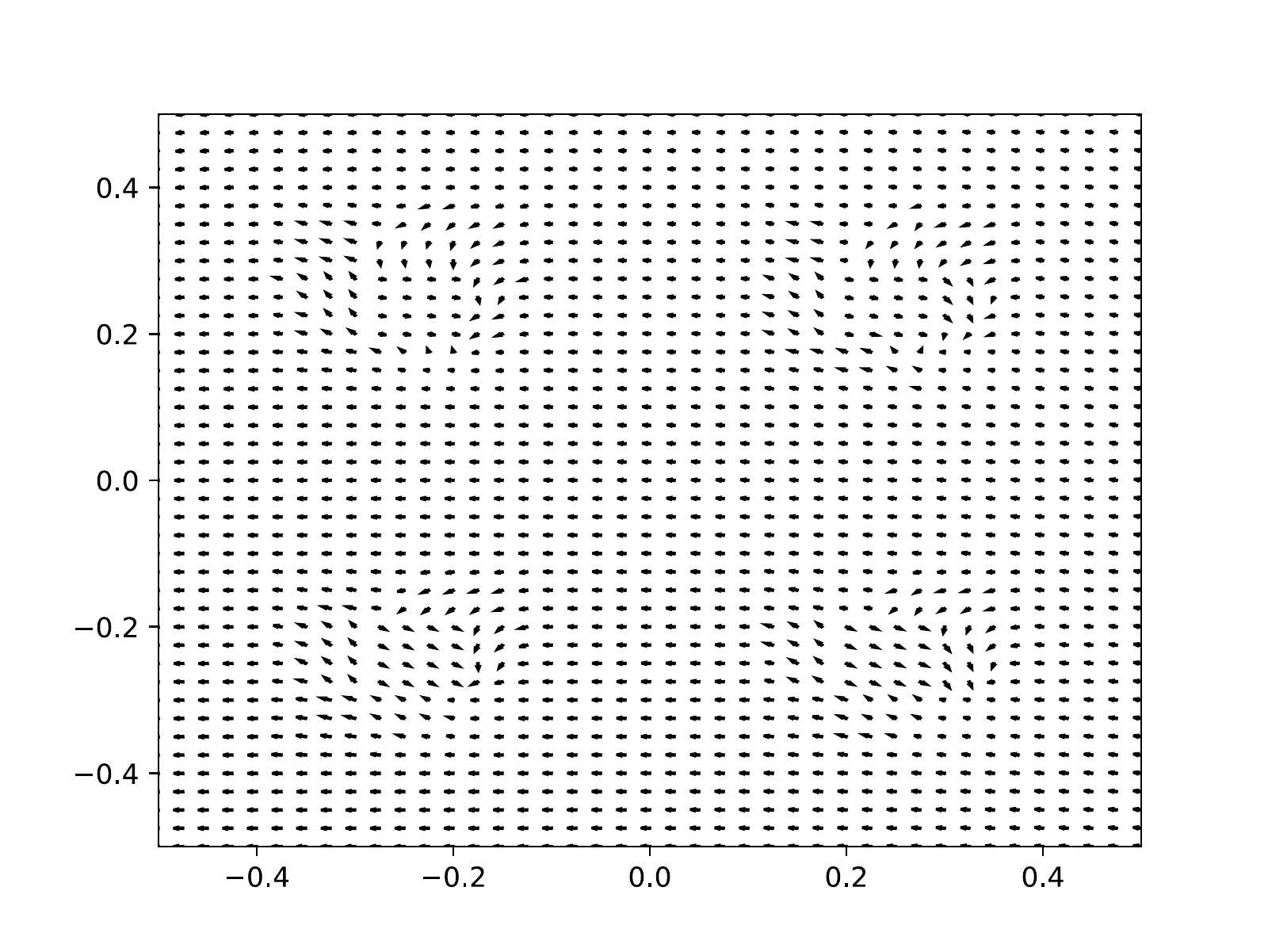}
        \caption{Generation 1}
    \end{subfigure}
    \begin{subfigure}{0.15\textwidth}
        \includegraphics[width=1.0\textwidth]{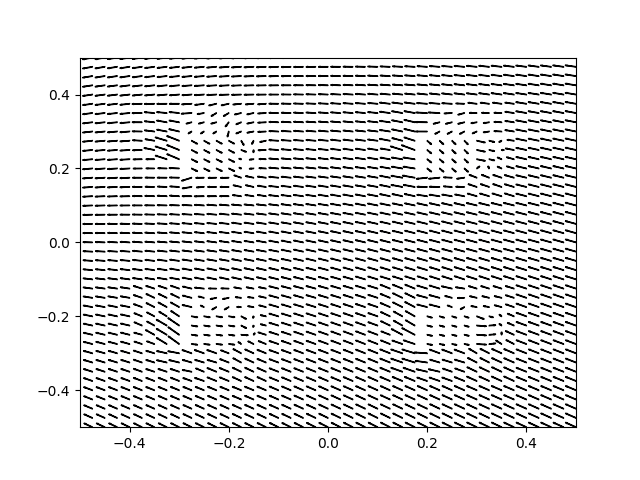}
        \caption{Generation 100}
    \end{subfigure}
    \begin{subfigure}{0.15\textwidth}
        \includegraphics[width=1.0\textwidth]{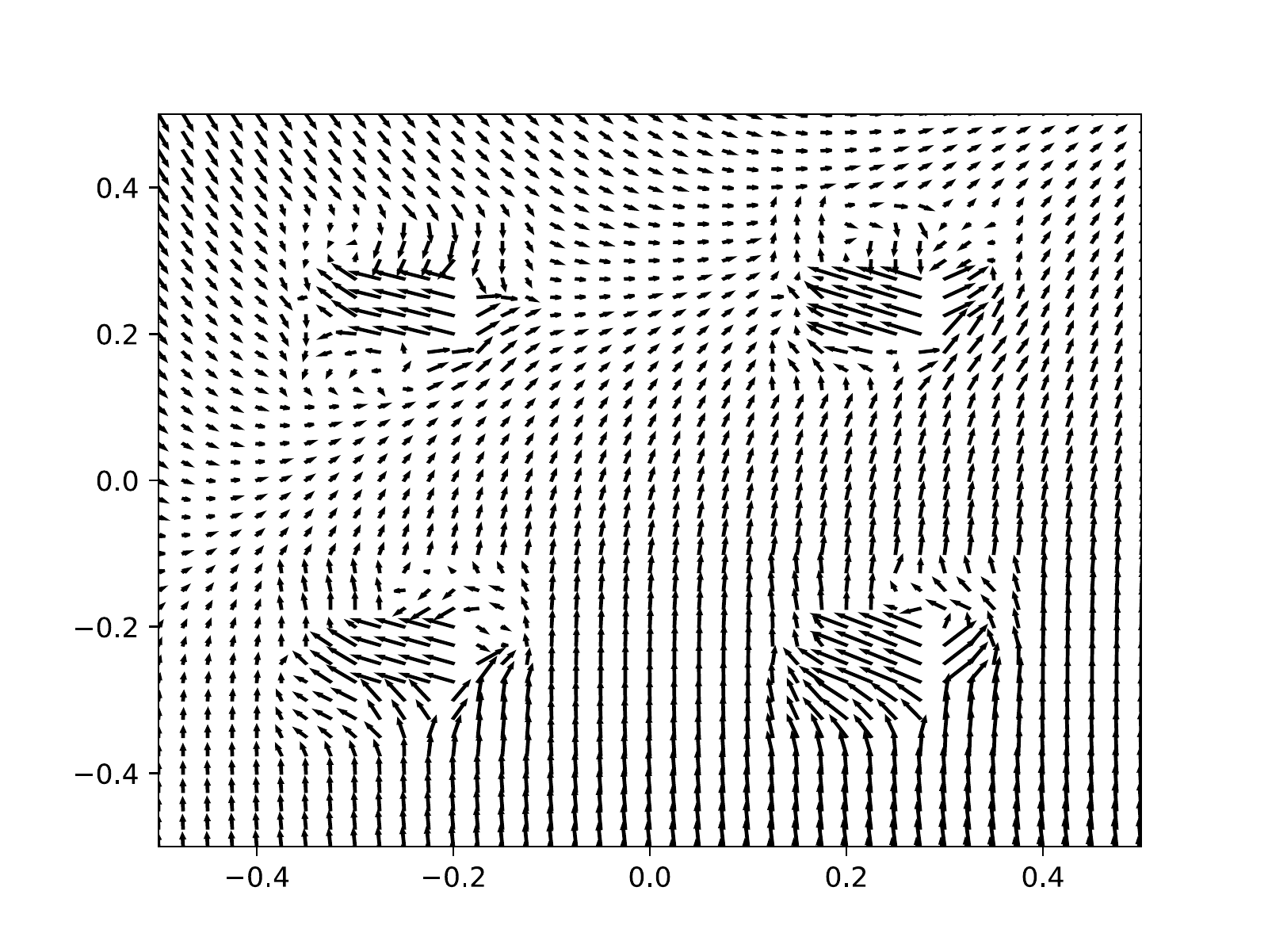}
    \caption{Generation 250}
    \end{subfigure}
    \caption{Modulated instinctual actions $\vec{a}_i^*$ mapped over the corresponding states $s$. \normalfont 
As the individual instinct improves, the pattern of actions around the hazard zones appears to strongly deviate from the surrounding action pattern. 
        }
\label{arrow_map}
\end{figure}

To gain a deeper understanding of the function of the evolved instinct network, we plot the average instinct suppression signal pattern based on the location on the 2D-navigation environment plane (Fig.~\ref{arrow_map}). As the instinct evolves, the average modulated instinctual action that is added to the modulated policy action increases in magnitude. The final evolved instinct  changes the direction of bias added to the policy in states close to the hazardous zones, preventing the policy from entering the hazardous  zones.

\begin{figure}
\centering
\begin{subfigure}{0.17\textwidth}
    \includegraphics[width=1.0\textwidth]{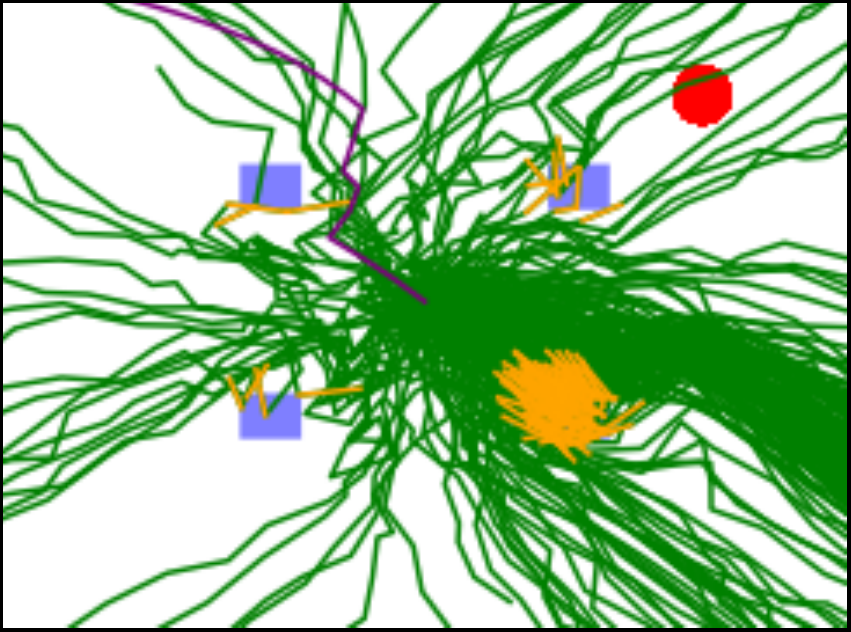}
    \caption{Instinct turned off}
\end{subfigure}
\begin{subfigure}{0.17\textwidth}
    \includegraphics[width=1.0\textwidth]{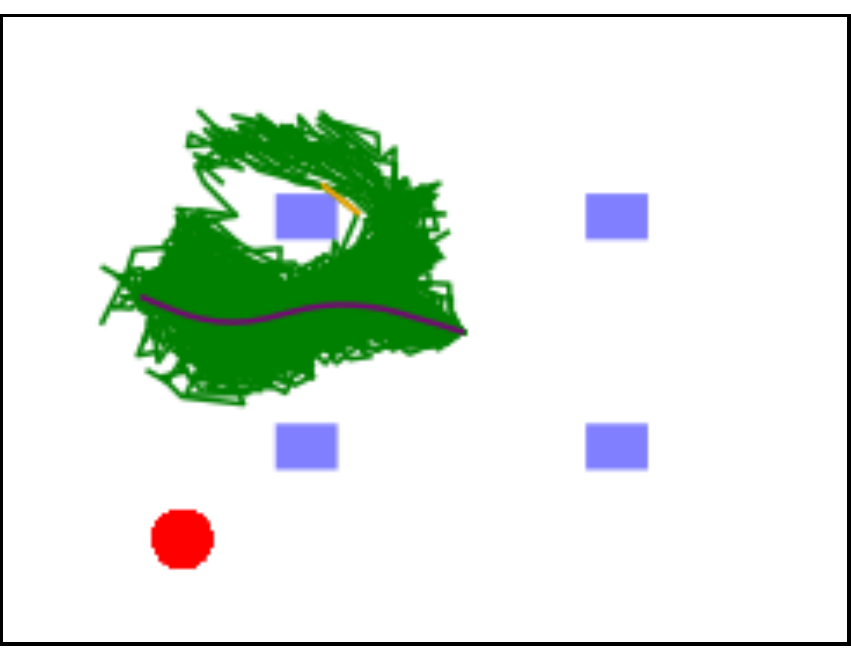}
    \caption{Only instinct}
\end{subfigure}
\caption{Exploration trajectories in the 2D navigation environment with the ablated model. \normalfont (a) shows the exploration trajectories of the evolved model shown in Fig.~\ref{fig:trajectories}a with the instinct module turned off. The orange lines mark when the agent stepped inside hazardous areas. The trajectories of the model with instinct turned on but the policy network randomly initialized are shown in (b).
}
\label{fig:ablation}
\end{figure}

\subsection{Ablation studies}
The exploration trajectories of the best performing model evolved with MLIN with evolved instinct network turned off are shown in Fig.~\ref{fig:ablation}a. Not surprisingly, removing the instinct reduces the ability of the model to reach the goal  (Table~\ref{table:fitness}) and to avoid the hazardous zones (Table~\ref{table:violations}). Fig.~\ref{fig:ablation}b shows the stochastic trajectories of an evolved MLIN model in which the initial evolved parameters of the policy network are replaced with random weights and Gaussian action noise $\sigma=0.05$. The instinct network by itself is able to steer the  agent with the random policy away from hazards. 

The advantage of the instinct module is that by changing its weights over the longer (evolutionary) time span the transformation that it produces over the action space can safely regulate the randomness of actions to avoid dangerous states. The main result is that MLIN produces a network that can better adapt to different goals than any of the other methods (Table~\ref{table:fitness}) while doing so in a safe way (Table~\ref{table:violations}). 

\begin{table}[]
\centering
\caption{\normalfont Average fitness over 20 repetitions.}
\begin{tabular}{ll}
\hline
Method & Avg. dist. fitness \\ \hline
 MLIN       &     -3.9 $\pm$ 1.5   \\
Meta-learning \emph{without} instincts  &     -11.3 $\pm$ 10.4   \\
Pure RL  &     -13.3 $\pm$ 2.4   \\
MLIN (removed instinct)  &     -8.2 $\pm$ 0.37   \\
MLIN (randomly init policy)  &      -13.7$\pm$ 2.2   \\
\end{tabular}
\label{table:fitness}
\end{table}

\begin{table}[]
\centering
\caption{\normalfont \normalfont Average hazardous zone violations over 20 repetitions.  MLIN  is the only approach able to avoid dangerous zones while also closely approaching the targets.}
\begin{tabular}{ll}
\hline
Method  & Avg. violations \\ \hline
MLIN       &     0.05 $\pm$ 0.2   \\
Meta-learning \emph{without} instincts  &     0.03 $\pm$ 0.16   \\
Pure RL  &     75.9 $\pm$ 48.3   \\
MLIN (removed instinct)  &     8.6 $\pm$ 2.0   \\
MLIN (randomly init policy)  &     33.4 $\pm$ 6.6   \\
\end{tabular}
\label{table:violations}
\end{table}

\section{Discussion and Future Work}
Safety in deep RL is a prerequisite condition for someday applying these methods in the real world. Here, we demonstrate that a slowly changing instinct component that can regulate the noisy actions of a policy network during the exploration can avoid hazardous areas while consistently adapting to specific goals in a simple navigation environment. 
Interestingly, the solution the meta-trained model finds is not to completely suppress the actions from the policy network, but to redirect them by changing the direction of the average instinctual action (Fig.~\ref{arrow_map}).  

Adapting the setup presented here to more challenging tasks is an important next step. One such environment is the recently published OpenAI Safety Gym benchmark \citep{raybenchmarking}, which contains multiple different tasks such as reaching goals, pushing objects toward goals, and avoiding static and moving dangers during learning. 

\section*{Acknowledgments}
This work was supported by the Lifelong Learning Machines program from DARPA/MTO under Contract No. FA8750-18-C-0103.Any opinions, findings and conclusions or recommendations ex-pressed in this material are those of the author(s) and do not necessarily reflect the views of DARPA

\footnotesize
\bibliographystyle{apalike}
\bibliography{main}

\end{document}